\documentclass[10pt,twocolumn,letterpaper]{article}

\usepackage{cvpr}              

\usepackage{graphicx}
\usepackage{amsmath}
\usepackage{amssymb}
\usepackage{booktabs}
\usepackage{multirow}
\usepackage[accsupp]{axessibility}
\usepackage{algorithm}
\usepackage{algpseudocode}
\usepackage[pagebackref,breaklinks,colorlinks]{hyperref}

\usepackage[table]{xcolor}
\usepackage[capitalize]{cleveref}
\crefname{section}{Sec.}{Secs.}
\Crefname{section}{Section}{Sections}
\Crefname{table}{Table}{Tables}
\crefname{table}{Tab.}{Tabs.}
\definecolor{vision}{rgb}{0.92,0.96,1.0}
\definecolor{state}{rgb}{0.89,0.98,0.89}
\newcommand{\state}{\rowcolor{state}}
\newcommand{\vision}{\rowcolor{vision}}

\def\cvprPaperID{3339} 
\def\confName{CVPR}
\def\confYear{2023}

\begin{document}

\title{
PartManip: Learning Cross-Category Generalizable Part Manipulation Policy \\ from Point Cloud Observations
}

\author{
Haoran Geng \textsuperscript{1,2*} \quad 
Ziming Li \textsuperscript{1,2*} \quad
Yiran Geng \textsuperscript{1,2} \quad 
Jiayi Chen \textsuperscript{1,3} \quad
Hao Dong \textsuperscript{1,2} \quad
He Wang\textsuperscript{1,2†} \\
\textsuperscript{1}CFCS, Peking University \quad
\textsuperscript{2}School of EECS, Peking University \\
\textsuperscript{3}Beijing Academy of Artificial Intelligence  
}

\twocolumn[{%
\renewcommand\twocolumn[1][]{#1}%
\maketitle
\begin{center}
    \centering
    \captionsetup{type=figure}
    \vspace{-1em}
    \includegraphics[width=1\linewidth]{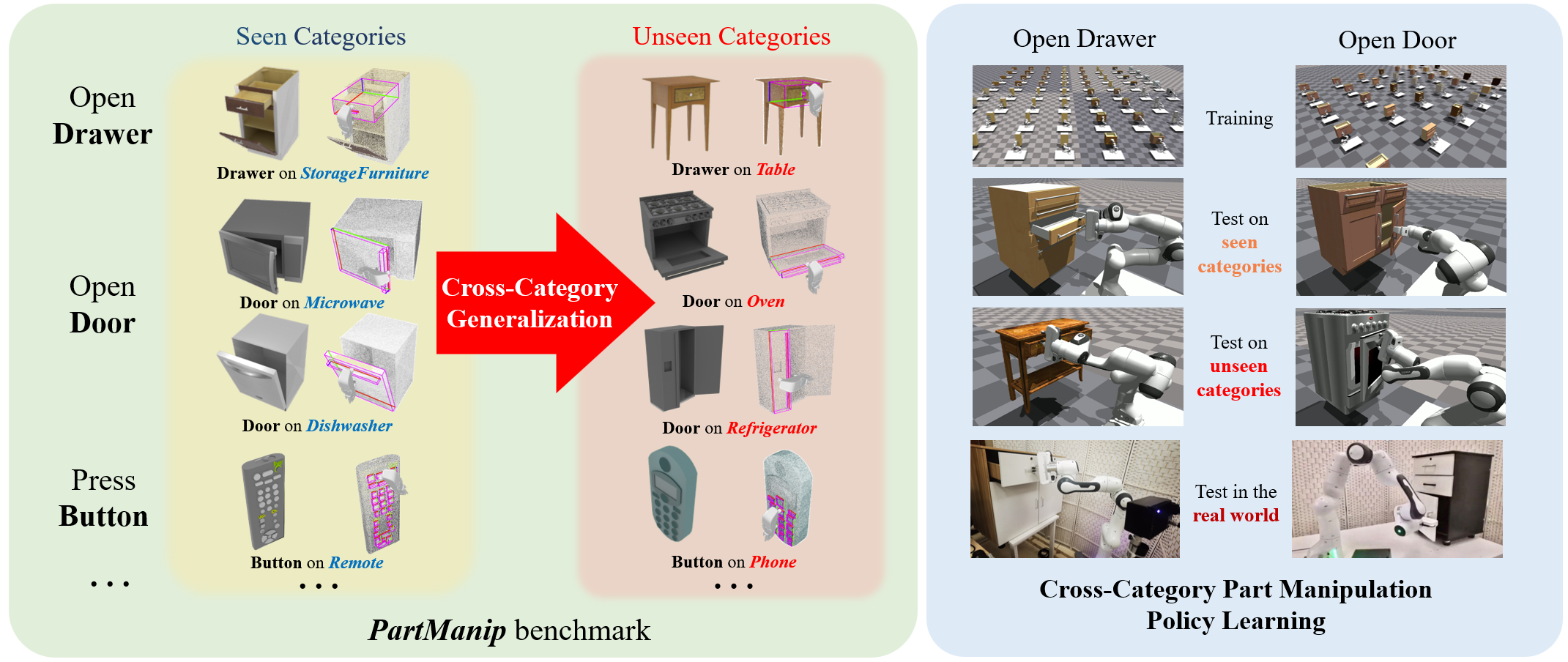}
    \vspace{-1em}
    \captionof{figure}{\textbf{Overview.}   We introduce a large-scale cross-category part manipulation benchmark \textbf{\textit{PartManip}} with diverse object datasets, realistic settings, and rich annotations. We propose a generalizable vision-based policy learning strategy and boost the performance of part-based object manipulation by a large margin, which can generalize to unseen object categories and novel objects in the real world.}
    \label{fig:teaser}
\end{center}
}]

\renewcommand{\thefootnote}{\fnsymbol{footnote}}
\footnotetext[1]{Equal contribution.}
\footnotetext[2]{Corresponding author: \href{mailto:hewang@pku.edu.cn}{hewang@pku.edu.cn}.}

\begin{abstract}

   Learning a generalizable object manipulation policy is vital for an embodied agent to work in complex real-world scenes. 
   Parts, as the shared components in different object categories, have the potential to increase the generalization ability of the manipulation policy and achieve cross-category object manipulation. 
   In this work, we build the first large-scale,  part-based cross-category object manipulation benchmark, \textbf{PartManip}, which is composed of 11 object categories, 494 objects, and 1432 tasks in 6 task classes. Compared to previous work, our benchmark is also more diverse and realistic, \textit{i.e.}, having more objects and using sparse-view point cloud as input without oracle information like part segmentation. 
   To tackle the difficulties of vision-based policy learning, we first train a state-based expert with our proposed part-based canonicalization and part-aware rewards, and then distill the knowledge to a vision-based student. We also find an expressive backbone is essential to overcome the large diversity of different objects. For cross-category generalization, we introduce domain adversarial learning for domain-invariant feature extraction. Extensive experiments in simulation show that our learned policy can outperform other methods by a large margin, especially on unseen object categories. We also demonstrate our method can successfully manipulate novel objects in the real world. Our benchmark has been released in \href{https://pku-epic.github.io/PartManip}{https://pku-epic.github.io/PartManip}.
\end{abstract}

\section{Introduction}

We as humans are capable of manipulating objects in a wide range of scenarios with ease and adaptability. For building general-purpose intelligent robots that can work in unconstrained real-world environments, it is thus important to equip them with generalizable object manipulation skills. Towards this goal, recent advances in deep learning and reinforcement learning have led to the development of some generalist agents such as GATO~\cite{reed2022generalist} and SayCan~\cite{ahn2022can}, however, their manipulation skills are limited to a set of known instances and fail to generalize to novel object instances. ManiSkill\cite{mu2021maniskill} proposes the first benchmark for learning category-level object manipulation, \textit{e.g.}, learn open drawers on tens of drawer sets and test on held-out ones. However, this generalization is limited within different instances from one object category, thus falling short to reach human-level adaptability. The most recent progress is shown in GAPartNet~\cite{1812.05276}, which defines several classes of generalizable and actionable parts (GAParts), e.g. handles, buttons, doors, that can be found across many different object categories but in similar ways. For these GAPart classes, the paper then finds a way to consistently define GAPart pose across object categories and devise heuristics to manipulate those parts, e.g., pull handles to open drawers, based on part poses. As a pioneering work, GAPartNet points to a promising way to perform cross-category object manipulation but leave the manipulation policy learning unsolved. 

In this work, we thus propose the first large-scale, part-based cross-category object manipulation benchmark, \textbf{\textit{PartManip}}, built upon GAPartNet. Our cross-category benchmark requires agents to learn skills such as opening a door on storage furniture and generalizing to other object categories such as an oven or safe, which presents a great challenge for policy learning to overcome the huge geometry and appearance gaps among object categories.

Furthermore, our benchmark is more realistic and diverse. We use partial point clouds as input without any additional oracle information like part segmentation masks in the previous benchmark ManiSkill~\cite{mu2021maniskill, gu2023maniskill2}, making our setting very close to real-world applications. Our benchmark also has much more objects than ManiSkill. We selected around 500 object assets with more than 1400 parts from GAPartNet\cite{geng2022gapartnet} and designed six classes of cross-category manipulation tasks in simulation. Thanks to the rich annotation provided in GAPartNet, we can define part-based dense rewards to ease policy learning.

Due to the difficulty presented by our realistic and diverse cross-category setting, we find that directly using state-of-the-art reinforcement learning (RL) algorithms to learn a vision-based policy does not perform well. Ideally, we wish the vision backbone to extract informative geometric and task-aware representations, which can facilitate the actor to take correct actions. However, the policy gradient, in this case, would be very noisy and thus hinder the vision backbone from learning, given the huge sampling space. To mitigate this problem, we propose a two-stage training framework: first train a state-based expert that can access oracle part pose information using reinforcement learning, and then distill the expert policy to a vision-based student that only takes realistic inputs.

For state-based expert policy learning, we propose a novel part-based pose canonicalization method that transforms all state information into the part coordinate frame, which can significantly reduce task variations and ease learning. In addition, we devise several part-aware reward functions that can access the pose of the part under interaction, providing a more accurate guide to achieve the manipulation objective. In combination, these techniques greatly improve policy training on diverse instances from different categories as well as a generalization to unseen object instances and categories. 

For the vision-based student policy learning, we first introduce a 3D Sparse UNet-based backbone~\cite{graham2017submanifold} to handle diverse objects, yielding much more expressivity than PointNet. To tackle the generalization issue, we thus propose to learn domain-invariant (category-independent) features via introducing an augmentation strategy and a domain adversarial training strategy~\cite{ganin2016domain,ganin2015unsupervised,li2018domain}. These two strategies can alleviate the problem of overfitting and greatly boost the performance on unseen object instances and even categories. Finally, we propose a DAgger~\cite{ross2011reduction} + behavior clone strategy to carefully distill the expert policy to the student and thus maintain the high performance of the expert.

Through extensive experiments in simulation, we validate our design choices and demonstrate that our approach outperforms previous methods by a significant margin, especially for unseen object categories (more than 20\% of the success rate in OpenDoor and OpenDrawer tasks). We also show real-world experiments.

\vspace{-2mm}
\section{Related Work\label{sec:relwork}}
 \vspace{-2mm}
\begin{table*}[t]
\centering
\setlength\tabcolsep{2pt} 
\renewcommand{\arraystretch}{0.95}
\begin{tabular}{c|c|c|c|c|c|c}
\hline
               & Generalization Level  & \# of Door Ins. & \# of Door Cat. & \# of Drawer Ins. & \# of Drawer Cat. & Realistic Input\\ 
    \hline
ManiSkill 1\&2   &    Category-level      & 82  & 1 & 70 & 1 & \\ 
\textbf{Ours}    & \textbf{Cross-category-level} & \textbf{503} & \textbf{7} & \textbf{399}& \textbf{3} & \checkmark \\ \hline
\end{tabular}
\caption{\textbf{Comparision with ManiSkill 1\&2~\cite{mu2021maniskill, gu2023maniskill2}.} Realistic input indicates whether to need oracle part segmentation masks.}
\label{table:BenchCom}
\end{table*}
\subsection{Learning Generalizable Manipulation Skills}
 
Generalization is crucial yet challenging for robot application.
Many works\cite{fang2020graspnet, sundermeyer2021contact, gou2021rgb, gao2021kpam, xu2021adagrasp} combine supervise learning and motion planning to learn generalizable skills, \textit{e.g.} grasping. However, these methods often require special architecture design for each task and may be unsuitable for complex manipulation tasks. Reinforcement learning (RL) has the potential to solve more complex task\cite{rajeswaran2017dapg, akkaya2019solving}, but the generalization ability of RL is an unsolved problem\cite{kirk2021generalization_survey,ghosh2021generalization}. 
To facilitate the research of generalizable manipulation skills, ManiSkill \cite{mu2021maniskill} proposes the first category-level object manipulation benchmark in simulation.  \cite{shen2022learning} leverages imitation learning to learn complex generalizable manipulation skills from the demonstration. However, their demonstrations are collected by RL trained on every single instance, which requires a lot of effort to tune. In contrast, our experts can be directly trained on each object category.

\subsection{Vision-based Policy Learning}
 
A lot of efforts are made to study how to learn policy from visual input\cite{zhang2015towards
,kalashnikov2018scalable, srinivas2020curl, yarats2021improving, stooke2021decoupling, geng2022end, xu2023unidexgrasp}.
Some works\cite{seo2022reinforcement, radosavovic2022real, seo2022masked} use a pre-trained vision model and freeze the backbone to ease the optimization. Some \cite{wu2022learning, wu2022vatmart} leverage multi-stage training. The most related work to us is \cite{chen2022system}, which also trains a state-based expert and then distills to a vision-based policy, but the task is quite different.
 
\subsection{3D Articulated Object Manipulation}
 
Manipulating articulated objects is an important and open research area due to the complexity of different objects' geometry and physical properties. Previous work has proposed some benchmark \cite{urakami2019doorgym, mu2021maniskill} but the diversity is limited. As for the methods, \cite{peterson2000high, jain2009pulling, burget2013whole} explored motion planning, and \cite{mo2021where2act,wu2022vatmart,wang2022adaafford,zhao2022dualafford, geng2022end} leverages visual affordance learning. Other works \cite{eisner2022flowbot3d, xu2022universal} design special representations for articulated object manipulation and can generalize to a novel object category, but their methods are only suitable for the suction gripper. 

\vspace{-1mm}
\section{PartManip Benchmark}
\vspace{-1mm}
\label{sec:dataset_def}

By utilizing the definition of the generalizable and actionable part (GAPart) presented in GAPartNet~\cite{geng2022gapartnet}, we build a benchmark for a comprehensive evaluation of the cross-category generalization policy learning approaches. GAParts are some kinds of parts that have similar geometry and similar interaction strategy across different object categories. For example, the handle on tables is often similar to those on safes, so we can regard the handle as a GAPart. The nature of GAPart ensures a general way for manipulation regardless of the object category, making it possible for cross-category generalization. We thus expect the manipulation policy trained on some object categories can generalize to other unseen object categories, and build the first benchmark for cross-category generalizable part manipulation policy learning.

Furthermore, our benchmark is more diverse and realistic than previous robotic manipulation benchmarks \cite{mu2021maniskill}. Diversity indicates that we have more object instances and categories, as shown in Table \ref{table:BenchCom}. Realism indicates that our observation space has less oracle information (\textit{i.e.}, part segmentation masks) than ManiSkill \cite{mu2021maniskill} as discussed in Sec. \ref{sec:obs_space}, and thus is more acquirable in the real world. 

We use Isaac Gym \cite{makoviychuk2021isaac} as our simulator and most experiments are done in simulation. In the following, we introduce our cross-category part-based object manipulation tasks in detail.

\begin{table*}[]
\setlength\tabcolsep{0.7pt} 
\centering
\resizebox{1\textwidth}{!}{
\begin{tabular}{c|ccccccc|ccc|cccc|cccccccc}
\hline
 & \multicolumn{7}{c|}{\textbf{Open}/\textbf{CloseDoor}} & \multicolumn{3}{c|}{\textbf{Open}/\textbf{CloseDrawer}}& \multicolumn{4}{c|}{\textbf{PressButton}}& \multicolumn{8}{c}{\textbf{GraspHandle}}\\ \hline

 &Sto. &Mic. &Dis. &Ove. &Ref. &Tab. &Saf. &Sto. &Tra. &Tab. &Rem. &Was. &Mic. &Pho. &Tra. &Tab. &Sto. &Mic. &Saf. &Ove. &Dis. &Ref.\\
\hline
Training Set &338&4&21&-&-&-&-   &238&8 &-  &178&30&7&-  &8&36&40&3&1&-&-&-\\
Val-Intra Set &60&-&3&-&-&-&-     &53 &4 &-  &47 &11&1&-  &4&16&20&1&-&-&-&-\\
Val-Inter Set &-&-&-&20&43&13&1   &-  &- &96 &-  &- &-&40 &-&- &- &-&-&20&24&43\\ \hline
Total        &398&4&24&20&43&13&1&291&12&96 &225&41&8&40 &12&52 &60 &4&1&20&24&43\\ \hline
\end{tabular}}
\caption{\textbf{Task Statistics and Split.} Sto. = StorageFurniture, 
Mic. = Microwave, 
Dis. = Dishwasher, 
Ove. = Oven, 
Ref. = Refrigerator, 
Tab. = Table , 
Saf. = Safe, 
Tra. = Trashcan, 
Rem. = Remote , 
Was. = WashingMachine, 
Pho. = Phone. 
Note that all the numbers are the task number instead of the object number, \ie, one object can have multiple parts and thus multiple tasks.
}
\label{tab:assets}
\end{table*}

\begin{figure*}[t]
\centering
\vspace{-0.3cm}
\includegraphics[width=1\linewidth]{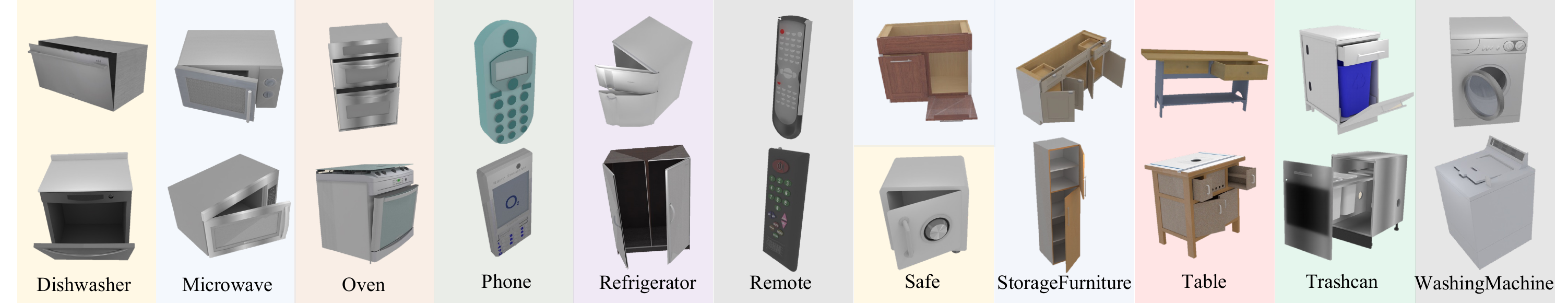}
\caption{\textbf{Object Assets Visualization.}  The object geometry and appearance are very different, especially in different object categories, which presents a great challenge for our \textbf{\textit{PartManip}} benchmark.}
\label{fig:gallery}
\end{figure*}

\begin{figure*}[t]
\centering
\vspace{-3mm}
\includegraphics[width=1\linewidth]{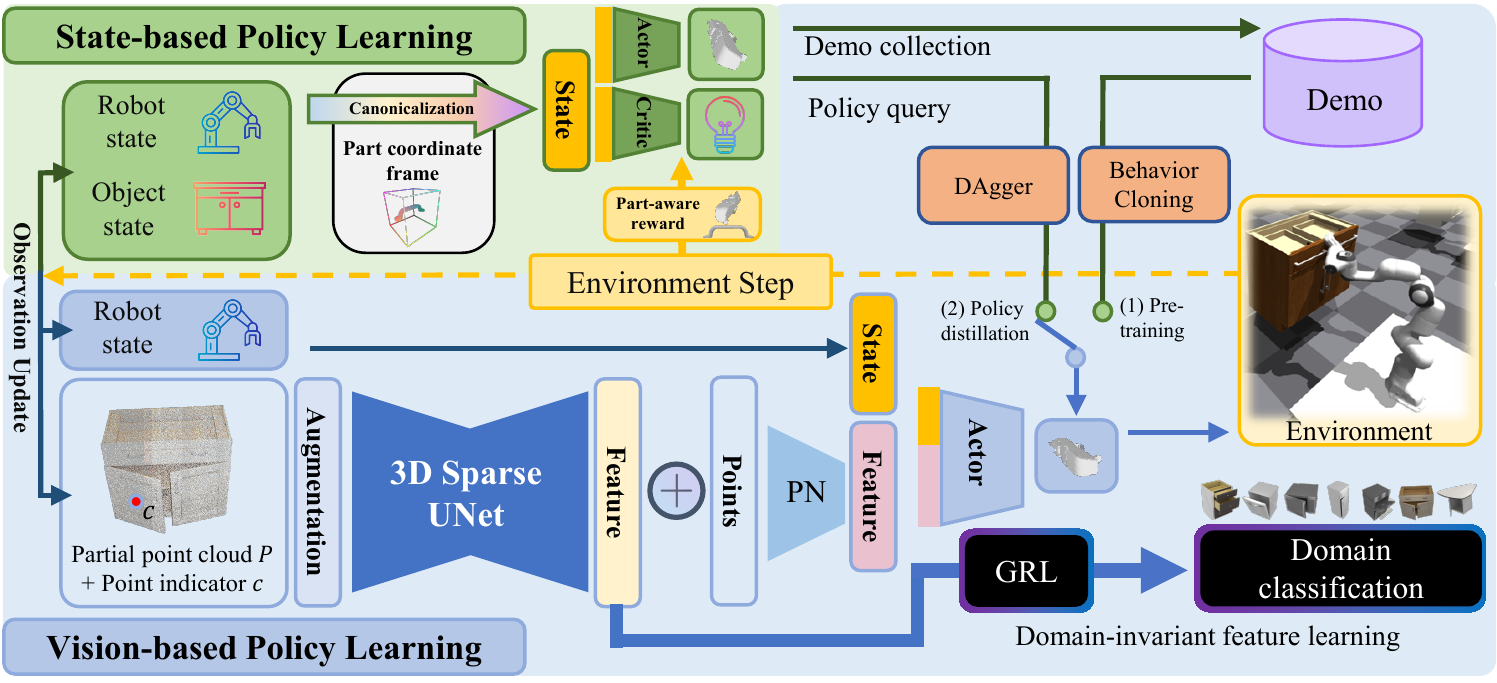}
\vspace{-3mm}
\caption{\textbf{Our Pipeline.} We first train state-based expert policy using our proposed canonicalization to the part coordinate frame and the part-aware reward. We then use the learned expert to collect demonstrations for pre-training the vision-based policy by behavior cloning. After pre-training, we train the vision-based policy to imitate the state-based expert policy using DAgger. We also introduce several point cloud augmentation techniques to boost the generalization ability. For the vision backbone, we introduce 3D Sparse-UNet which has a large expression capability. Furthermore, we introduced an extra domain adversarial learning module for better cross-category generalization.}
\label{fig:pipeline}
\end{figure*}

\subsection{Task Formulation}
 
We have six classes of tasks: \textbf{OpenDoor}, \textbf{OpenDrawer}, 
\textbf{CloseDoor}, \textbf{CloseDrawer}, \textbf{PressButton} and \textbf{GraspHandle}. Although \textbf{OpenDoor} and \textbf{OpenDrawer} require grasping the handle first, \textbf{GraspHandle} differs from them because it contains another GAPart \textit{lid} with more object categories. Like traditional RL tasks, our task can be formulated as a Partially Observable Markov Decision Process (POMDP), because the true environment state $s_t$ is not fully observable (especially for objects) in each timestep $t$. Given current partial observation $o_t$, a policy $\pi$ needs to predict an action $a_t$ to control the robot. After applying the action  $a_t$, the next observation $o_{t+1}$ and the reward $r_{t+1}$ will be given by the environment, which can be used to train the policy. 
The final goal of the policy is to reach the success state (see supp. for more) in $T$ steps (we set episode length $T=200$).
 
\subsection{Object Assets}
 
For each task, our object assets contain part-centered articulated objects, which are selected from the GAPartNet dataset \cite{geng2022gapartnet}. GAPartNet dataset is a large-scale articulated object dataset with GAPart definitions and rich part annotations, including semantics and poses. 
The original GAPart classes contain \textit{Round Fixed Handle, Slider Lid, Slider Button, Hinge Knob, Line Fixed Handle, Slider Drawer, Hinge Lid} and \textit{Hinge Door}. But in this paper, we simplify them to only five classes: \textit{handle, button, door, drawer} and \textit{lid}. Each \textit{door/drawer/lid} has a \textit{handle} on it for manipulation. Figure \ref{fig:gallery} shows some visualization of our object assets, and Table \ref{tab:assets} shows the statistics and split in our benchmark.

\subsection{State Space and Action Space}
 
The environment state $s$ of each task is composed of the robot joint angles $qp \in\mathbb{R}^{11}$, joint velocity $qv \in\mathbb{R}^{11}$, gripper coordinate $ x \in \mathbb{R}^3$, gripper rotation $r \in SO(3)$, and the target GAParts' bounding boxes $b_\text{GAParts} \in \mathbb{R}^{3\times8\times n}$, where $n$ is the number of the target GAParts for the manipulation task. Specifically, there are two GAParts \textit{handle} and \textit{door/drawer/lid} for all the tasks in our benchmark except for \textbf{PressButton}, which only has one GAPart \textit{button}. 

For the action space, we use the Franka Panda arm and a parallel gripper as the end-effector. The action $a_t$ is formulated as the gripper's target 6 Dof pose at each timestep $t$ for all tasks. It is then converted to the target joint angles by inverse kinematics (IK), and used to control the robot arm by position control.
\vspace{-1mm}
\subsection{Observation Space}
\vspace{-1mm}
\label{sec:obs_space}
Because the true states of GAParts are often not observable in a realist setting, our acquirable input for policy learning is the partial observation $o_t$, which consists of the robot states $(qp, qv, x, r)_t$, the colored point clouds $P_t \in \mathbb{R}^{6\times4096}$, and a fixed point indicator $c \in \mathbb{R}^3$ indicating the center of the target GAPart. 

In our paper, we use a multi-camera setting  in simulation to alleviate occlusion. The colored point clouds $P_t$ are back-projected from 3-view RGBD images, assuming the camera poses are known. One camera is set above the object facing down, and the other two are set on each side of the object facing the objects. See We our supp. for the experiments with the single-camera setting.

The point indicator $c$ is necessary for disambiguation because some objects have multiple actionable parts (\textit{e.g.}, double-door refrigerator) and we need a way to specify which part we want the robot to manipulate. Specifically, for those tasks that have both \textit{handle} and \textit{door/drawer/lid}, we only give the center of the target \textit{door/drawer/lid} at timestep $0$ as the point indicator $c$. The policy has to find and manipulate the corresponding \textit{handle} by only point clouds. Compared to ManiSkill 1\&2 \cite{mu2021maniskill}, which uses the oracle segmentation mask for disambiguation, our point indicator $c$ is easier to acquire in the real world and poses a greater challenge for policy learning. Note that the point indicator $c$ does not update over time, so it does not leak extra information about the part's motion.

\vspace{-1mm}
\subsection{Part Pose-aware Reward Design}
\vspace{-1mm}
We design dense rewards for policy training because we can access the ground truth (GT) environment state $s$ in the simulator. Note that the state $s$ is not available during testing, which makes it hard for direct motion planning. Inspired by ManiSkill\cite{mu2021maniskill}, we designed a general reward for all tasks. The reward has 4 components:

\textbf{Rotation Reward}. GAParts such as \textit{handle} and \textit{button} can only be manipulated at certain angles. 
The reward $R_\text{rot}=\cos(\overrightarrow{a_p} \cdot \overrightarrow{a_{r}})$, in which ${{a}_p}\in\mathbb{R}^{3}$ is the y-axis of the part bounding box and ${a}_{r}\in\mathbb{R}^{3}$ is the y-axis of robot gripper. When the grippper is vertical to the face of the target part, the reward reach the maximum value.

\textbf{Distance Reward}. The reward is the negative value of the distance from the center of two tips on gripper ${c}_\text{g}\in\mathbb{R}^{3}$ to the handle center ${c}_\text{h}\in\mathbb{R}^{3}$. We also add a discount factor ${d_f} \sim {d_f}(b_{GAParts})$ to reduce the distance reward after the target part has been removed. The intuition is that, for example, when the target door has been opened large enough, our humans won’t tightly grasp the handle in a rigid hand pose. So, we won't encourage the gripper to grasp the handle as strictly as before either. Finally, $R_\text{dist}= - d_f \left\| {c}_\text{g} - {c}_\text{h} \right\|_2$. 
\textbf{Part Moving Reward}. The reward is computed by the movement of the target GAPart, encouraging the robot to move the part to the success state. For example, for \textbf{OpenDoor}, it's the radian degree of the open angle; for \textbf{OpenDrawer}, it's the open length. We formulate it as 
${R}_\text{pose}$.

\textbf{Tips closure Reward}. The reward encourages tips to close for stable grasping. It's computed by the distance of two fingertips on grippers. We formulate it as 
${R}_\text{tips}$.

The whole reward can be written as:
$${
\lambda_r{R}_\text{rot} 
+ \lambda_d R_\text{dist}
+ \lambda_p{R}_\text{pose}
+ \lambda_t{R}_\text{tips}}
$$
We use one set of hyper-parameters for each task and the specific value is shown in the appendix.

\section{Method \label{sec:method}}

%
 Our \textbf{\textit{PartManip}} benchmark is very difficult and thus simple methods can fail dramatically. Our benchmark requires the learned policy not only to manipulate multiple objects from partial visual observation but also to generalize the manipulation skill to unseen object categories. Directly applying state-of-the-art RL algorithms (\textit{e.g.}, PPO\cite{schulman2017proximal}) on our benchmark to learn a vision-based policy cannot perform well, possibly due to the unstable RL training process. Therefore, we need specific designs to tackle the difficulty.
 
 We start with expert policy learning using oracle environment state as input. Thanks to the rich part annotations in our benchmark, we propose a novel part-canonicalized strategy for policy learning in \cref{sec: state_rl}, which greatly improves the performance and generalization ability. 

Given the state-based expert, we then introduce a knowledge distillation strategy to learn a vision-based student policy in \cref{sec: vision_rl}, which takes input from the partial observation $o$ instead of the oracle environment state $s$. Our State-to-Vision Distillation training strategy also enables visual observation augmentation to boost the generalization ability. We thus introduce our point cloud augmentation technique here.

Additionally, due to the wide variety of objects and the inherent difficulty in processing visual input, it is crucial to use a backbone with large expression capability. To this end, we propose a Sparse Unet-based backbone architecture\cite{graham2017submanifold} for policy learning in \cref{sec: backbone}, which offers superior feature extraction performance. 

Last but not least, to overcome the challenges presented by cross-category task settings, we present a domain adversarial training strategy\cite{ganin2016domain,ganin2015unsupervised,li2018domain,geng2022gapartnet} in \cref{sec: domain} for our visual feature learning. This strategy improves the generalizability of the policy, particularly on novel object categories.


\subsection{Part-Canonicalized State-Based RL}
\label{sec: state_rl}

Compared to the partial visual observation $o$, using the oracle environment state $s$ as input can greatly reduce the difficulty of policy learning because the network will not be distracted by visual feature extraction and can focus on learning manipulation skills. So we start with state-based expert policy learning using a state-of-the-art RL algorithm Proximal Policy Optimization (PPO). 

PPO is a popular on-policy RL algorithm, which proposes a few first-order tricks  to stabilize the training process. The stability is ensured by keeping the new policy close to the old one by gradient clipping or KL penalty. PPO trains an actor and a critic, and the objective to maximize can be formulated as
  
 \begin{equation}
 \mathcal{L} = \mathbb{E} \left[ \frac{\pi_\theta(a|s)}{\pi_{\theta_{old}}(a|s)} \hat{A} - \epsilon KL[\pi_{\theta_{old}}(\cdot |s), \pi_\theta(\cdot |s)] \right] 
 \end{equation}
 in which $\theta$ is the network parameters, $\epsilon$ is a hyper-parameter and $\hat{A}$ is the estimated advantage using GAE~\cite{schulman2015high}. Thanks to the GPU-parallel data sampling in Isaac Gym\cite{makoviychuk2021isaac}, PPO can converge fast and stably, so we choose it as our baseline and develop on it. 

In our scenario, we find that the relative pose between the robot gripper and the target GAPart ( \textit{e.g., handle/button}) is more critical than the absolute pose. Take the task of \textbf{OpenDoor} as an example, if the relative poses between two \textit{handles} and two grippers are similar, we can use a similar strategy to manipulate them. It doesn't matter whether these \textit{handles} are on a microwave or a oven. It also doesn't matter what the absolute poses of the gripper and \textit{handles} are. 

Therefore, we propose to transform all the coordinate-dependent states from world space to the target GAPart's canonical space. 
Inspired by the normalized coordinate space (NOCS~\cite{wang2019normalized}) for category-level object pose estimation, we define the canonical space of each GAPart category similarly using the part bounding box annotations provided by GAPartNet~\cite{1812.05276}. For each GAPart category, we choose the center of the part bounding box as the origin point, and three axes align to the three vertical edges of the bounding box. 
Here the scale is ignored because it can change the size of the robot gripper and bring potential harm. For each frame $t$, we can use the target GAPart's pose (in the format of bounding box $b_{GAPart}^t$) to transform all the coordinate-dependent states into the GAPart's canonical space.

This part-canonicalized strategy can transform all the absolute poses into relative poses to the target GAPart, and thus reduce the variance of input. In our experiments in \cref{sec:exp}, we show that this strategy can greatly improve the generalization ability of the learned policy, and achieve a high success rate even on unseen object categories.

\subsection{Augmented State-to-Vision Policy Distillation}

\label{sec: vision_rl}
\noindent\textbf{DAgger-based State-to-Vision Distillation.} The learned expert cannot be directly used in a realistic setting because in the real world the oracle environment state $s$ is often unknown and we can only acquire the partial observation $o$. However, we can leverage the state-based expert knowledge to ease the learning process of the vision-based student. This technique is often called knowledge distillation~\cite{gou2021knowledge}. 

One popular and simple method for knowledge distillation is Behavior Cloning (BC). BC requires the expert to collect an offline dataset of observation-action pairs with success trajectories.  The objective of BC is to minimize the difference between the action label and the student policy output. Although it sounds reasonable and straightforward, it can suffer from accumulating error over time because the observation distribution of the student policy is different from the one of the offline dataset.

To overcome the problem of accumulating error, we use DAgger \cite{ross2011reduction, chen2022system} for knowledge distillation. DAgger is an online imitation learning algorithm, which can directly train on the observed distribution of the student policy thanks to online interaction. The objective of DAgger is similar to BC except for the data distribution, which can be written as:
 
\begin{equation}
\mathcal{L}_{DA} = \frac{1}{\left\vert {D}_{\pi_\theta} \right\vert} 
\sum_{o,s \in {D}_{\pi_\theta}} 
\left\| {\pi}_\text{expert} (s) - {\pi}_\text{student}(o) \right\|_2 
\end{equation}

in which ${D}_{\pi_\theta}$ is the online interacted data sampled from student policy during training. 

\noindent\textbf{Pretraining with BC.} One potential issue we observed in DAgger is that a randomly initialized policy may perform poorly in the first few iterations, stepping out of the expert's input distribution. Since the expert policy has not encountered those out-of-distribution states, it may not know how to act correctly to finish the task. So using the output of expert policy in such states for supervision can slow down the training process of students, leading to worse performance. 

To address this issue, we use the learned expert to collect offline demonstrations and pre-train the student policy with BC. Although pure offline BC cannot be superior to pure online DAgger, BC is a good choice for initialization and can boost the further DAgger training process.
Using this pre-training technique along with DAgger, we employ two different ways to make full use of the expert knowledge.

\noindent\textbf{Observation Augmentations.} Thanks to our state-to-vision distillation method, the student policy will not suffer from the noisy gradient in RL and thus can make full use of its network capability. To improve the generalization ability of the learned policy, a promising way is to enlarge the training distribution of the object geometry and appearance. Therefore, we propose to add augmentations on point clouds during the DAgger training process. These augmentations mainly include point cloud jittering and color replacement. Please see the appendix for more details.

\subsection{3D Sparse Unet-based Backbone}

\label{sec: backbone}
In order to effectively process complex and diverse visual input for cross-category generalization, it is essential to have an expressive backbone. To this end, we utilize the widely-used backbone, 3D Sparse-UNet\cite{graham2017submanifold}, which has a stronger expression capacity compared to PointNet\cite{qi2017pointnet} and PointNet++\cite{qi2017pointnet++}. Sparse-UNet is often employed in state-of-the-art methods such as PointGroup\cite{jiang2020pointgroup} and SoftGroup\cite{vu2022softgroup}. However, a significant drawback of Sparse-UNet is its slow running speed. In order to address this issue, we introduce several algorithms (such as batch voxelization for point clouds and high parallelization of sparse convolution) to optimize the implementation of Sparse-UNet. These modifications greatly speed up the forward and backward processes of the network, resulting in a runtime that is over 1.4 times faster than the best existing implementation. Additional details can be found in the appendix.

As illustrated in Fig. \ref{fig:pipeline}, the partial colored point cloud $P\in \mathbb{R}^{N\times 6}$ is used as input to Sparse-UNet, which extracts per-point features $F \in \mathbb{R}^{N\times C}$. These per-point features are then processed by a small PointNet\cite{qi2017pointnet} and concatenated with the robot states to serve as input to the actor.

\subsection{Domain-invariant Feature Learning}
 
\label{sec: domain}
Inspired by \cite{ganin2016domain,ganin2015unsupervised,li2018domain,geng2022gapartnet},  we introduce the Gradient Reverse Layer (GRL) and a domain classifier for domain-invariant feature learning, which improves the generalization ability across object categories, especially for unseen categories. The domain classifier takes the extracted visual features from the backbone as input to distinguish domains, \ie object categories in our setting, while the GRL negates the gradient and encourages the backbone to fool the classifier and thus extract domain-invariant features. For a target part with mask $M_i$, we first query its feature $F_{M_i}$ from the whole feature map $F$. Then a classifier $\mathcal{D}$ with a small Sparse-UNet backbone and two MLPs takes it as input and classifies the domain label. The loss can be written as :
\begin{equation}
\mathcal{L}_\text{adv} = \mathcal{L}_{cls}(\mathcal{D}(F_{M_i}), y^\text{cate.}_i)
\end{equation}
\noindent where $y^\text{cate.}_i$ is the ground truth domain label (\ie, category label).
\label{sec: pre_joint_train}

\section{Experiments}
\label{sec:exp}
 
\subsection{Evaluation Settings and Main Results}

Because our \textit{\textbf{PartManip}} benchmark emphasizes the experiment settings to be realistic, all the algorithms are evaluated using partial observation $o$ as input except those experts in \cref{table:Ablation}. We use the task success rate as our main evaluation metric. To reduce the evaluation noise, we  conduct each experiment 3 times using random seeds and report the mean performance as well as the variance. See appendix for more training details.

The main results of our method are shown in \cref{table:MainResult}. We can see that \textbf{CloseDoor} and \textbf{CloseDrawer} are relatively easy and our method can achieve a high success rate, even on unseen object categories. The other four tasks are more challenging and the performance drops on the unseen categories, especially for \textbf{OpenDoor} and \textbf{OpenDrawer}.
\subsection{Comparison with Baselines}
 
To demonstrate the effective of our method, we conduct several relative baselines on our \textit{\textbf{PartManip}} benchmark. These baselines include a popular RL algorithm PPO \cite{schulman2017proximal}, an affordance-based interaction method W2A~\cite{mo2021where2act}, a novel vision-based policy training method ILAD~\cite{wu2022learning}, and winners \cite{shen2022learning,pan2022silverbullet3d,wu2022minimalist,dubois2022improving} 
in ManiSkill Challenge~\cite{mu2021maniskill}.
Due to the page limit, we only show the results of two representative tasks, \ie, \textbf{OpenDoor} and \textbf{OpenDrawer}, in our main paper. 
See the appendix for the performance of the other four tasks.

As shown in \cref{table:Baselines}, we can see that although most baselines can work to some degree on the training set, the performance can drop dramatically on intra- and especially inter-category validation set. In contrast, our method can perform consistently better on all evaluation set without intense performance drop, which indicates the great generalization ability of our method.

\begin{table}[t]
\centering
\setlength\tabcolsep{2pt} 
\renewcommand{\arraystretch}{0.95}
\begin{tabular}{l|ccc}
\hline
         Success rate (\%)     & Training Set & Val-Intra Set & Val-Inter Set \\ \hline
CloseDoor    & 88.7$\pm$1.0 & 88.4$\pm$2.9 & 87.0$\pm$1.6\\ 
CloseDrawer  & 99.6$\pm$0.6 & 97.9$\pm$2.1 & 98.6$\pm$1.2 \\ 
OpenDoor    & {68.4$\pm$1.1}&{57.2$\pm$0.4}&{49.1$\pm$1.5} \\ 
OpenDrawer  & {82.3$\pm$2.1}&{78.7$\pm$2.0}&{54.7$\pm$4.2} \\ 
PressButton & 89.6$\pm$2.9& 79.6$\pm$4.2 & 66.6$\pm$4.2\\ 
GraspHandle  & 79.8$\pm$2.4& 70.0$\pm$2.4& 56.4$\pm$2.9 \\ \hline
\end{tabular}
\caption{\textbf{Our Main Results on \textit{PartManip} benchmark.} 
}
\label{table:MainResult}

\end{table}

\begin{table}[t]
\centering
\setlength\tabcolsep{2pt} 
\renewcommand{\arraystretch}{0.95}
\resizebox{.5\textwidth}{!}{
\begin{tabular}{l|ccc|ccc}
\hline
\multirow{2}{*}{Success rate (\%)} & \multicolumn{3}{c|}{OpenDoor}& \multicolumn{3}{c}{OpenDrawer}   \\ 
 & Training & Val-Intra & Val-Inter & Training & Val-Intra & Val-Inter  \\
\hline \multirow{1}{*}{\begin{tabular}[c]{@{}c@{}}PPO\cite{schulman2017proximal}\end{tabular}}
&4.5$\pm$3.8&4.9$\pm$3.5&0.2$\pm$0.2&8.9$\pm$2.8&11.3$\pm$2.8&3.3$\pm$1.6\\
\multirow{1}{*}{\begin{tabular}[c]{@{}c@{}}ILAD\cite{wu2022learning}\end{tabular}}
&13.3$\pm$4.9&6.3$\pm$2.5&5.0$\pm$4.1&18.7$\pm$3.6&18.3$\pm$2.9&3.3$\pm$2.9\\
\multirow{1}{*}{\begin{tabular}[c]{@{}c@{}}Where2act\cite{mo2021where2act}\end{tabular}}
&25.4$\pm$0.1&23.4$\pm$0.0&15.2$\pm$0.1&39.6$\pm$0.2&37.2$\pm$0.2&20.5$\pm$0.1\\
SilverBullet3D~\cite{pan2022silverbullet3d}&54.6$\pm$2.5&49.9$\pm$1.0&26.9$\pm$2.2&77.7$\pm$3.3&60.0$\pm$2.0&31.2$\pm$5.1\\
\textit{Shen et. al}~\cite{shen2022learning}&1.5$\pm$0.6&0.3$\pm$0.6&2.3$\pm$4.0&9.7$\pm$0.5&18.0$\pm$2.2&2.7$\pm$1.9\\

\textit{Wu et. al}~\cite{wu2022minimalist}&45.9$\pm$2.3&34.1$\pm$3.8&17.8$\pm$1.4&70.5$\pm$2.2&53.3$\pm$3.3&28.5$\pm$2.4\\
\textit{Dubois et. al}~\cite{dubois2022improving}&35.4$\pm$4.4&25.1$\pm$2.1&1.3$\pm$1.0&61.4$\pm$4.3&38.3$\pm$2.0&2.7$\pm$1.6\\
         
\multirow{1}{*}{\begin{tabular}[c]{@{}c@{}}\textbf{Ours}\end{tabular}}
& \textbf{68.4$\pm$1.1}&\textbf{57.2$\pm$0.4}&\textbf{49.1$\pm$1.5}&\textbf{82.3$\pm$2.1}&\textbf{78.7$\pm$2.0}&\textbf{54.7$\pm$4.2}  \\ \hline
\end{tabular}}
\caption{\textbf{Comparison with Baselines.} }
\label{table:Baselines}
\vspace{-2mm}
\end{table}

\begin{table*}[t]
\centering
\setlength\tabcolsep{2pt} 
\renewcommand{\arraystretch}{0.95}
\resizebox{.99\textwidth}{!}{
\begin{tabular}{c|cccccc|ccc|ccc}
\hline
  \multirow{2}{*}{Success rate (\%)} & \multirow{2}{*}{Canon} &\multirow{2}{*}{DAgger} & \multirow{2}{*}{Augm}  & \multirow{2}{*}{S-Unet} & \multirow{2}{*}{Pretrain} & \multirow{2}{*}{DomAdv} & \multicolumn{3}{c|}{Opening Door}& \multicolumn{3}{c}{Opening Drawer}   \\ 
 
& && &&& & Training & Val-Intra & Val-Inter & Training & Val-Intra & Val-Inter \\ 
 \hline \state 
State-based 
 &          &          &          &          &          &&67.8±3.4&50.2±1.9& 23.4±3.9& 71.5±2.1& 62.5±2.3& 37.5±5.2  \\
 \state Expert &\checkmark&          &          &          &          &&\textbf{82.2$\pm$0.2}& \textbf{62.5$\pm$2.6}& \textbf{50.7$\pm$4.1}& \textbf{92.7$\pm$0.9} &\textbf{88.1$\pm$1.0}& \textbf{63.4$\pm$2.4}  \\   \hline \vision
             &          &          &          &          &          &&4.5$\pm$3.8&4.9$\pm$3.5&0.2$\pm$0.2&8.9$\pm$2.8&11.3$\pm$2.8&3.3$\pm$1.6  \\ \vision
           &          &          &          &\checkmark&          &&0.8$\pm$0.5&0.4$\pm$0.2&0.0$\pm$0.0&5.9$\pm$2.3&3.9$\pm$0.6&1.0$\pm$0.2 \\ \vision
            &          &\checkmark&          &          &          &&60.3$\pm$0.7&49.2$\pm$1.1&31.5$\pm$2.9&70.9$\pm$0.6 & 62.0$\pm$1.1 & 42.7$\pm$1.8  \\\vision Vision-based  
             &          &\checkmark&          &\checkmark&          &&66.8$\pm$2.7&50.2$\pm$1.7&28.8$\pm$2.1&77.4$\pm$2.7 & 61.9$\pm$3.0 & 36.4$\pm$3.3  \\\vision Student  
            &          &\checkmark&\checkmark&          &           &&60.0$\pm$1.7&54.4$\pm$2.3&40.2$\pm$3.9&69.7$\pm$2.4
            &69.8$\pm$2.5&49.0$\pm$2.1  \\ \vision 
            &          &\checkmark&\checkmark&\checkmark&          &&65.5$\pm$1.5&55.9$\pm$2.7&41.7$\pm$2.5&74.6$\pm$3.4&63.8$\pm$4.7&49.1$\pm$3.4  \\ \vision  
             &          &\checkmark&\checkmark&          &\checkmark&&61.1$\pm$3.3&55.0$\pm$1.2&37.8$\pm$2.9&71.9$\pm$3.3&72.2$\pm$3.5&50.3$\pm$2.6  \\ \vision
            &          &\checkmark&\checkmark&\checkmark&\checkmark&&\textbf{71.2$\pm$1.8}&57.0$\pm$0.7&37.2$\pm$1.2&82.0$\pm$3.3&73.8$\pm$2.9&48.8$\pm$4.5  \\ \vision
          &          &\checkmark&\checkmark&\checkmark&\checkmark&\checkmark&{68.4$\pm$1.1}&\textbf{57.2$\pm$0.4}&\textbf{49.1$\pm$1.5}&\textbf{82.3$\pm$2.1}&\textbf{78.7$\pm$2.0}&\textbf{54.7$\pm$4.2}  \\  \hline
\end{tabular}}
\caption{\textbf{Ablation Study.} 
Canon = Part pose canonicalization for the input states; Augm = Augmentation; S-Unet = Sparse-UNet; Pretrain = Pretraining with expert demonstration; DomAdv = Domain adversarial learning.}
\label{table:Ablation}
\end{table*}

\subsection{Ablation Study}
The ablation studies are shown in \cref{table:Ablation}.
For state-based policy learning, our proposed part-based canonicalization can significantly improve performance on all three evaluation sets, especially for the inter-category validation set. Take \textbf{OpenDoor} as an example, the policy trained with part-based canonicalization can outperform the one without by 14.4\%, 12.3\%, and 27.3\% respectively on three sets. 

For vision-based policy learning, we analyze each component below (row number is counted from the visual policy):

1) (Row 1-4) Directly using the RL algorithm PPO to learn the policy performs poorly. We deduce that the RL gradient is noisy and can ruin the learning of a vision backbone, especially for a large network (\eg, Sparse-UNet). DAgger can greatly alleviate this problem and perform well on the training set, but it also suffers from overfitting and thus is lack of strong cross-category generalization ability.

3) (Row 3-6) Sparse-UNet backbone has a better expressive capacity but may overfit to the training set. Using augmentation can alleviate the overfit, achieving much better generalization ability.

5) (Row 5-8) Pretraining using the demo collected by the expert can provide a better initialization for the student, ease the  problem of being out of expert distribution at the beginning of DAgger, and improve performance.

6) (Row 8,9) Our domain adversarial learning can further boost the performance, especially on previously unseen instances or categories. We reason that our adversarial training strategy helps domain-invariant feature extraction and thus helps robotics policy learning.

\subsection{Real-World Experiments}
 
\begin{figure}[h]
\centering

\includegraphics[width=1\linewidth]{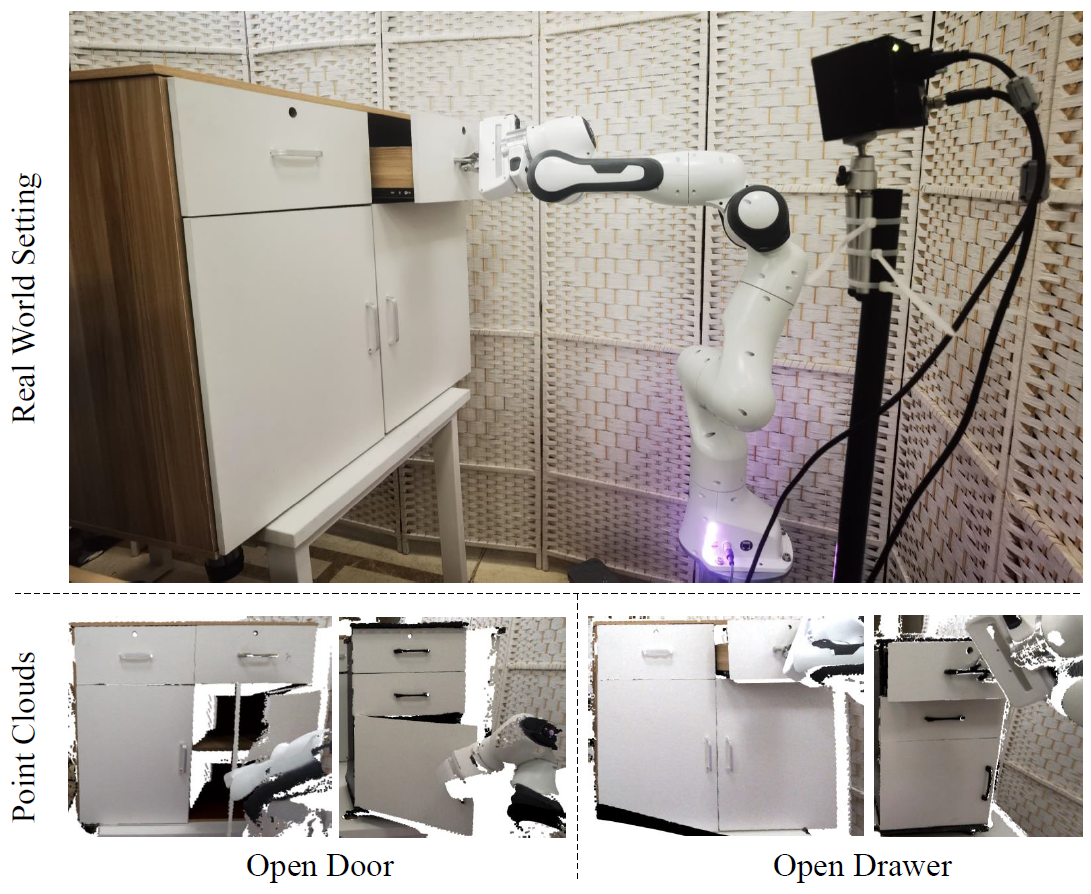}

\caption{\textbf{Real World Experiment.}}
\label{realworld}
\vspace{-3mm}
\end{figure}

Finally, we validate our method in the real world, as shown in Fig. \ref{realworld}. We use a single RGB-D camera (Okulo) to capture
visual observation and a Franka Emika robotic arm for manipulation.
We try to directly apply the learned policy in simulation to the real world but find the success rate is much lower, due to the huge sim2real gap of the point cloud and the physics. Therefore, to minimize the sim2real gap, we use a digital twin system~\cite{XIA2021210,geng2022end} for the real-world experiment. The input of our method included: 1) point cloud from the simulator and 2) agent state from the real world. We then apply the output action of our policy to the robotic arm both in the simulator and the real world. Note that the testing object is unseen during policy learning. 
Experiments show that our trained model can successfully transfer to the real world. See appendix for more details.

\section{Limitation}

Although our proposed method for cross-category object manipulation has already outperformed previous work by a large margin in simulation, there is still much room for improvement. It is worth studying how to further improve the performance on unseen object instances and categories. Another limitation can be the huge sim2real gap between the point cloud and the physics. We hope that our \textbf{\textit{PartManip}} benchmark can provide a good starting point for future research on generalizable manipulation policy learning.

\vspace{-2mm}
\section{Conclusion}
\vspace{-2mm}

In this work, we introduce a large-scale part-based cross-category object manipulation benchmark \textbf{\textit{PartManip}},  with six tasks in realistic settings. To tackle the challenging problem of the generalizable vision-based policy learning, we first introduce a carefully designed state-based part-aware expert learning method, and then a well-motivated state-to-vision distillation process, as well as a domain generalization technique to improve the cross-category generalization ability. Through extensive experiments in simulation, we show the superiority of our method over previous works. We also demonstrate the performance in the real world.

\vspace{-2mm}
\section{Acknowledgements}
\vspace{-2mm}
This work is supported in part by the National Key R\&D Program of China (2022ZD0114900).

{\small
\bibliographystyle{ieee_fullname}
\bibliography{egbib}
}

\newpage
\appendix

\section{More Details about Our Benchmark}
\subsection{Objects and Robot Assets}
We select 494 object instances from GAPartNet\cite{1812.05276} dataset. GApartNet provides large-scale articulated objects with rich annotations. For doors and drawers, our benchmark requires a handle on it. For the button, there is no constraint. Because there are too many buttons on some remote or phone, we randomly select a maximum of 5 buttons on each object. The total benchmark contains 494 objects and 1432 different target parts.

\subsection{Environment Settings}
We carefully set the environment parameters in Isaacgym.
To simulate real-world physics, we set the contact offset = 1e-3, which means that our gripper is harder to manipulate the part by scratching or rubbing the edge of the part. Also, we set a slight recovery force (= 0.1) to avoid the agent moving the part to a successful state by slightly touching it. The stiffness of the cabinet dof(slider joint for drawer and button, rotation joint for door) is set to 20. The damping of the cabinet dof is set to 200. The friction coefficient of cabinet dof is set to 5.

For robot controlling, we use pos control mode. This means that we need to input each joint angle of the Franka arm to the network at each step. We find that in our tasks, using pos control is easier for imitation learning.

\subsection{Reward Weight}
For each task, we tune the reward weight to train a human-like policy.  We list rotation weight $\lambda_r$ , handle distance weight $\lambda_d$, part moving weight $\lambda_p$, and tips closure weight $\lambda_t$ in Table \ref{table:task hyperparameter}.

\textbf{rotation weight}. For the closing task and pressing button task, we don't need the gripper to grasp the handle, so we set $\lambda_r$ = 0. For the other tasks, we set it to 0.2.

\textbf{handle distance weight}. The value of $\frac{\lambda_d}{\lambda_p}$ is important for opening task. If the value is too small, the policy wouldn't learn to open the part using the handle. If the value is too large, the policy wouldn't try to open the door, but only learn to grasp the handle. The $\lambda_{df}$ is set to non-zero for the opening task because after the part is opened, the gripper can move away from the handle.

\textbf{part moving weight}. For the grasping task, we don't require the agent to manipulate part, so we set $\lambda_p$ = 0. 

\textbf{tip closure weight}. For the opening and closing tasks, we focus on using the handle to finish the task but don't require the final grasp pose of the gripper. So we set $\lambda_t$ = 0.

\subsection{Initialization and Success Criteria}
\label{sec: Initialization and Success Criteria}
For each task, we initialize the gripper at a certain distance (\ie, 50cm) away from the center of the target part, facing the object. The initialization of objects and the success criteria for each task are shown below: 

\textbf{OpenDoor}: The door should be opened to more than 30 degrees from the initial closed state.

\textbf{OpenDrawer}: The drawer should be opened to more than 20\% of the maximum opening length from the initial closed state.

\textbf{CloseDoor}: The door should be closed to less than 1 degree from the initial opening angles of 45 degrees. 

\textbf{CloseDrawer}: The drawer should be closed from the initial opening length of 30 cm to less than 1 cm.

\textbf{PressButton}: The button should be pressed for more than 50\% of the maximum pressing depth.

\textbf{GraspHandle}:
The robot should close two tips to less than a threshold from different sides of the handle, while the center of the two tips is inside the handle bounding box. 


\section{More Details about Our Method}
Our pseudo code is shown in algorithm \ref{alg:expert} and algorithm \ref{alg:student}.  

For state-based policy training, we update $E$ epochs in one step. In one epoch, the state and actions buffer $D_{\pi_{\theta}}$ is divided in $B$ minibatch to compute one gradient step. So after sampling once, the network update $E * B$ time.

For state-to-vision distillation, we use point cloud augmentation. We use point cloud jittering with a distance of $0.1$ and a strong color augmentation, which changes the GAParts color to a random color during a specific episode. Although we randomly choose a color, we fix it during one episode. This technique works well and improves performance in the unseen category.

There is a potential problem for expert distillation. The output of the actor can be an arbitrary value in $\mathbb{R} ^ n$. Here $n$ is the output dimension of the actor. Because we use the pose control, and the joint angle is in the range $(-\pi, \pi)$. If the value $a_i$ in $i_{th}$ dimension is out of this range, it would shift to $a^ \prime _i$ satisfied $ a^ \prime _i = a_i + 2k\pi, k \in \mathbb{Z}$. Because of this, multiple outputs would correspond to one action in the simulator so the L2 loss of expert action and student action is not positively associated with the similarity between expert action and student action. To tackle this problem, we add an additional Tanh layer and scale the action to $(-\pi, \pi)$. We use the scaled action to compute dagger loss and update the network.

\begin{table*}[t]
\centering
\begin{tabular}{c|cccccc}
\hline
 & robot state & part bounding box & handle bounding box & Point Cloud & part mask & handle mask \\ \hline
Ours (state-based)   &  \checkmark &  \checkmark       &  \checkmark         &               &            &             \\ \hline
Where2Act \cite{mo2021where2act}         &  \checkmark  &                   &                     &  \checkmark  & \checkmark &   \\ 
VAT-Mart \cite{wu2022vatmart}         &    \checkmark&                   &                     &  \checkmark  & \checkmark &   \\ 
Maniskill \cite{mu2021maniskill}         &  \checkmark &                   &                     &  \checkmark  & \checkmark & \checkmark  \\ 

Ours (vision-based) &  \checkmark &                   &                     &  \checkmark  &            &   \\ \hline
\end{tabular}
\caption{\textbf{Comparison with Other Methods.}}
\end{table*}

\begin{table*}[t]
\centering
\setlength\tabcolsep{2pt} 
\renewcommand{\arraystretch}{0.95}
\resizebox{.95\textwidth}{!}{
\begin{tabular}{c|ccccccc}
\hline
name & opening door & opening drawer & closing door & closing drawer & pressing button & grasping handle \\ \hline
Train environment number & 363 & 246 & 363 & 246 & 215 & 88 \\ 
minibatchs & 2 & 3 & 2 & 3 & 2 & 2 \\
nsteps & 20 & 20 & 20 & 20 & 20 & 40 \\
$\lambda_r$ & 0.2 & 0.2 & 0 & 0 & 0 & 0.2 \\
$\lambda_d$ & 2 & 1.3 & 1 & 1 & 1 & 1\\
$\lambda_p$ & 1 & 1 & 1 & 1 & 100 & 0 \\
$\lambda_t$ & 0 & 0 & 0 & 0 & 10 & 1 \\ 
$\lambda_{df}$ & 1 & 2 & 0 & 0 & 0 & 0 \\ \hline
\end{tabular}}
\caption{\textbf{Task Specific Hyperparameters of State-based Policy Training}}
\label{table:task hyperparameter}
\end{table*}

\begin{algorithm}[t]
\caption{State-based Expert Training}\label{alg:expert}

\textbf{Input:}  robot states $S$, handle bounding box $b_{handle}$, part bounding box $b_{part}$, state $s = (S, b_{part}, b_{handle}) $, policy network $\theta_p$  ( \ie, actor $\theta_p^a$ and critic $\theta_p^c$),

\textbf{for} $t = 0,1,2\dots$ \textbf{do}

    \quad Transfer observation to canonical space
    
    \quad Sample trajectories $D_{\pi_{\theta}} = \{(s_i. a_i)\}_{i=1}^n$

    \quad \textbf{for} $e = 1,2,\dots, E$ \textbf{do} \Comment{PPO update}

    \quad \quad \textbf{for} $b = 1,2,\dots, B$ \textbf{do}
    
    \quad \quad \quad update policy network $\theta_p$, according to: $\mathcal{L}_{RL}$

Select highest success rate $\theta_p \to \theta_\text{expert}$

\textbf{while} {$\lvert D_{demo} \rvert \le $ buffer size} \textbf{do}

    \quad sample trajectory $t_i$ by $\theta_\text{expert}$ 
    
    \quad append $t_i$ to $D_{demo}$ 

\end{algorithm}

\begin{algorithm}[t]
\caption{Vision-based Student Training}\label{alg:student}
\textbf{Input:} partial point cloud $P \in \mathbb{R}^{N \times 3}$, robot states $s$, vision backbone $\theta_{b}$, policy network $\theta_p$  ( \ie, actor $\theta_p^a$ and critic $\theta_p^c$), expert policy $\theta_\text{expert}$,  demonstration buffer $D_\text{demo}$

\textbf{Pre-training:} update the vision backbone $\theta_b$ and actor MLPs $\theta_p^a$, according to $\mathcal{L}_\text{BC}$

\textbf{for} $t = 0,1,2\dots$ \textbf{do}

    \quad Sample trajectories $D_{\pi_{\theta}} = \{(s_i, o_i, a_i)\}_{i=1}^n$

    \quad \textbf{for} $e = 1,2,\dots, E$ \textbf{do}

    \quad \quad \textbf{for} $b = 1,2,\dots, B$ \textbf{do}
    
    \quad \quad \quad augment point cloud observation $o_i$ as $\mathcal{A}(o_i)$

    \quad \quad \quad update backbone $\theta_p$ and the actor of policy network  
    
    \quad \quad \quad $\theta_p^a$, according to: $\lambda_\text{DA}\mathcal{L}_\text{DA} + \lambda_\text{adv}\mathcal{L}_\text{adv}$

\end{algorithm}

\section{More Details about the Experiment Setting}

\subsection{Training Details}
We train our state-based policy on Nvidia GeForce RTX 2080Ti for 6 hours. For each task, we use all of the data in our dataset claimed in the paper. The PPO hyperparameter is shown in Table \ref{table:hyperparameter}.

For actor and critic networks, we use 3 hidden layers of MLP. The hidden layer dimension of the MLP is 512, 512, 64.  For vision-based policy, we use the Sparse-Unet backbone.  If the input point cloud has more than $N = 20,000$ points, we first downsample it to 20000 points using FPS (Farthest Point Sampling). Then we voxelize the input point cloud into a $100\times100\times100$ voxel grid. The backbone U-Net has an encoder and decoder, both with a depth of 6 (with channels of [16, 32, 48, 64, 80, 96, 112]) and outputs a $N \times K$ per-point feature $\mathbf{F}$ where $K = 16$.  We speed up the 3D Sparse UNet inference speed by introducing batch voxelization for point clouds and high parallelization of sparse convolution, thanks to the latest high-performance third-party code base like open3d and sparse 

Because PPO is an on-policy RL algorithm, for each $N$ step, we update the policy. To leverage the fast convergence of PPO, we want to update as frequently as we can. On the other side, due to the noisy gradient of RL, the batch that is used to compute the gradient should not be too small, which is equal to $T_{env} *N / M $. Here $T$ is the training environment number, and $M$ is the minibatch size. Empirically we find that the batch size near to 2000 is fine. For six tasks, due to the number of training data, we choose proper minibatch and nsteps. The task-specific hyperparameters are shown in Table  \ref{table:task hyperparameter}.

\begin{table}[H]
\centering
\setlength\tabcolsep{2pt} 
\renewcommand{\arraystretch}{0.95}
\begin{tabular}{c|c}
\hline
PPO params & value / type \\ \hline
learning rate & 3e-4 \\
optimizer   & Adam \\
gamma      & 0.99  \\
lambda     & 0.95  \\
desired kl & 0.01  \\
clip range & 0.1 \\
entropy coef & 0.01  \\
init noise std & 1 \\  \hline
\end{tabular}
\caption{\textbf{PPO Hyperparameters of Policy Training}}
\label{table:hyperparameter}
\end{table}

\subsection{More Details and Results of the Baselines }

\begin{table}[t]
\centering
\setlength\tabcolsep{2pt} 
\renewcommand{\arraystretch}{0.95}
\resizebox{.45\textwidth}{!}{
\begin{tabular}{c|c|ccc}
\hline
task & method & Training Set & ValIntra Set & ValInter Set \\ \hline
\multirow{4}{*}{Closing Door(\%)}&Where2act\cite{mo2021where2act} & 77.3$\pm$0.1&54.6$\pm$0.0 &51.5$\pm$0.2 \\
&PPO\cite{schulman2017proximal} & 35.5$\pm$1.1 &37.6$\pm$0.9 &15.4$\pm$0.5\\
&DAgger\cite{ross2011reduction} &84.5$\pm$2.5 &79.4$\pm$1.1 &69.9$\pm$2.3 \\
&Ours & \textbf{88.7$\pm$1.0} & \textbf{88.4$\pm$2.9} & \textbf{87.0$\pm$1.6} \\
\hline
\multirow{4}{*}{Closing Drawer(\%)}&Where2act\cite{mo2021where2act} &89.9$\pm$0.2 &90.5$\pm$0.1 &89.9$\pm$0.3 \\
&PPO\cite{schulman2017proximal} &69.9$\pm$5.9 &75.2$\pm$2.6 &59.3$\pm$2.1 \\
&DAgger\cite{ross2011reduction} &95.9$\pm$1.2 &97.3$\pm$1.1 &91.5$\pm$0.2 \\
&Ours & \textbf{99.6$\pm$0.6} & \textbf{97.9$\pm$2.1} & \textbf{98.6$\pm$1.2} \\
\hline
\multirow{4}{*}{Pushing Button(\%)}&Where2act\cite{mo2021where2act} & 15.5$\pm$0.2&16.2$\pm$0.1 & 19.3$\pm$0.3\\
&PPO\cite{schulman2017proximal} &25.5$\pm$0.2 &21.6$\pm$1.1 &7.9$\pm$5.5 \\
&DAgger\cite{ross2011reduction} &32.8$\pm$2.2 &\textbf{41.2$\pm$6.6} &29.8$\pm$1.2 \\
&Ours & \textbf{89.6$\pm$2.9}& \textbf{79.6$\pm$4.2} & \textbf{66.6$\pm$4.2}\\
\hline
\multirow{4}{*}{Grasping Handle(\%)}&Where2act\cite{mo2021where2act} & 27.7$\pm$0.1 &25.4$\pm$0.2 &13.9$\pm$0.3 \\
&PPO\cite{schulman2017proximal}& 15.7$\pm$2.2& 13.2$\pm$0.6&9.9$\pm$3.5 \\
&DAgger\cite{ross2011reduction} &45.6$\pm$2.2 &35.5$\pm$2.1 & 29.8$\pm$2.9\\
&Ours & \textbf{79.8$\pm$2.4}& \textbf{70.0$\pm$2.4}& \textbf{56.4$\pm$2.9}\\
\hline
\end{tabular}}
\caption{\textbf{More Results of Method Comparison and Baselines}}
\label{table:result}
\end{table}

For opening the door and drawer, thank the previous exploration, We compare our policy with many baselines. For \cite{shen2022learning,pan2022silverbullet3d,wu2022minimalist,dubois2022improving}, they focus on tasks like opening drawers and doors and we can modify their method to our OpenDoor and OpenDrawer tasks. And for the other four tasks, we also compare with some possible methods if they can be easily modified to fit our framework.
More results are shown in Table \ref{table:result}.

\textbf{PPO}\cite{schulman2017proximal}. We directly use the PPO algorithm to learn a vision-based policy to handle each task. The detailed PPO parameter and training strategy is the same as the state-based expert training in our method.

\textbf{Where2Act} \cite{mo2021where2act}. We input the part mask as an extra dimension in our task as a baseline, and others remain the same.
We modified the where2act interaction pipeline to finish our tasks. We use a similar pulling motion for the first three tasks and a pushing motion for the fourth task. Giving only a point to indicate the part to be interacted with makes it challenging for where2act to perform proper actions, especially for opening drawers and doors. We thus provide additional information (\ie, the handle center of the target door and drawer), and this method needs to select one point from the given points. Then, after motion direction selection, the action is performed to finish the task. We constrain $N_\text{w2a} = 10$ actions to finish these tasks.

\textbf{ILAD}\cite{wu2022learning}. Due to we have designed a dense reward in our task, we use our dense reward instead of their extra Q functions to compute the advantage in the third term of $g_{ILAD}$. The demonstrations are also collected by expert policy as the GAIL \cite{shen2022learning} baseline implementation. We don't input part 6D pose into the network as a fair comparison to our method.

\textbf{ManiSkill\cite{mu2021maniskill} Winners}, \ie, \textit{Shen et. al}~\cite{shen2022learning}, SilverBullet3D~\cite{pan2022silverbullet3d}, \textit{Wu et. al}~\cite{wu2022minimalist}, \textit{Dubois et. al}~\cite{dubois2022improving}. We follow the ManiSkill\cite{mu2021maniskill} settings and follow the corresponding policy learning strategy to learn. If the method needs collected demonstrations as input, we provide the demonstrations collected by the state-based expert.

For these baselines, we analyze that their performances are limited due to the distribution shift in behavior cloning, lacking vital information with realistic sensory observation input and noisy reinforcement learning gradient.

\subsection{More Results for a Single Camera Setting}
Here we provide more experiment results for a single camera setting. For OpenDoor, the performances are 36.7$\pm$3.3, 33.9$\pm$2.4, 22.6$\pm$3.0 in the training
set, Val-Intra set and Val-Inter set respectively. For OpenDrawer, the performances are 64.5$\pm$5.5, 60.2$\pm$4.4, 17.1$\pm$2.2 in the training
set, Val-Intra set and Val-Inter set respectively.

\subsection{Some Qualitative Results for the Failure Cases}
\begin{figure}
    \centering
    \includegraphics[width=1\columnwidth]{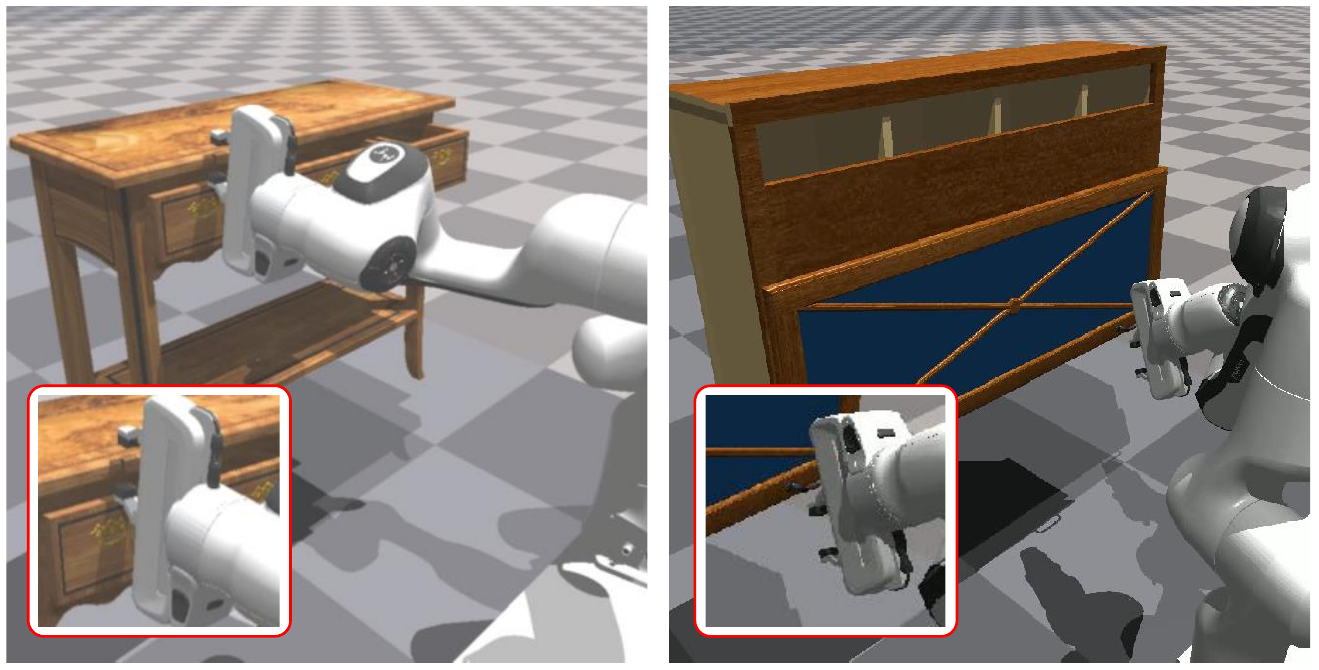}
    \caption{\textbf{Failure Cases}}
    \label{fig:failure}
\end{figure}
Here we provide some qualitative results for failure cases.
In Fig.~\ref{fig:failure}, we show two failure cases. For the left one, the gripper fails to identify the handle and grasps the wrong position due to the thin and flat handle shape (yellow, zoom in to see), while for the right one,  the door opening fails later for unstable grasping.

\section{Real Experiment}
We use the robot arm (FRANKA) to manipulate previously unseen real objects with only partial point cloud observations. A partial point cloud of the target object instance is acquired from the RGB-D camera (Okulo P1 ToF sensor in our experiments). To set up the interaction environment, we use aruco markers to calibrate the camera sensor and place the object and the robot arm in the proper positions, the same as the trained policy in the simulator. We also provide a point to indicate the part to interact with, just like we did in the simulator. During manipulation, we use the control API provided by the robot arm system to follow the trajectory (a sequence of joint angle and gripper position) and finish the tasks.

\end{document}